\documentclass[twoside,10pt]{article}

\usepackage{mlhc}

\newcommand{\mname}{\texttt{medGAN}\xspace}
\usepackage{bm,amsmath,amssymb}
\usepackage{xspace}
\usepackage{color}

\usepackage[toc,page]{appendix}
\usepackage{subcaption}
\usepackage{sidecap}
\sidecaptionvpos{figure}{c}

\usepackage{wrapfig}

\usepackage{url}

\newcommand{\yrcite}[1]{\citeyearpar{#1}}
\renewcommand{\cite}[1]{\citep{#1}}

\newcommand{\hide}[1]{}

\usepackage{booktabs}       % professional-quality tables
\usepackage{amsfonts}       % blackboard math symbols
\usepackage{nicefrac}       % compact symbols for 1/2, etc.
\usepackage{microtype}      % microtypography

\usepackage{enumitem}

\usepackage{bm}

\newcommand{\Wb}{\mathbf{W}}
\newcommand{\Xb}{\mathbf{X}}

\newcommand{\xb}{\mathbf{x}}
\newcommand{\yb}{\mathbf{y}}
\newcommand{\zb}{\mathbf{z}}

\usepackage{times}
\usepackage{graphicx} % more modern
\usepackage{natbib}
\usepackage{url}

\usepackage{algorithm}
\usepackage{algorithmic}

\usepackage{hyperref}

\begin{document}

\vspace{-6mm}

\title{Generating Multi-label Discrete Patient Records\\ using Generative Adversarial Networks}

\author{\name Edward Choi$^1$ \email mp2893@gatech.edu
%\addr Georgia Institute of Technology 
        \AND
        \name Siddharth Biswal$^1$ \email sbiswal7@gatech.edu
        %\addr Georgia Institute of Technology
        \AND
        \name Bradley Malin$^2$ \email bradley.malin@vanderbilt.edu 
        %\addr Vanderbilt University
        \AND
        \name Jon Duke$^1$ \email jon.duke@gatech.edu
        %\addr Georgia Institute of Technology
        \AND
        \name Walter F. Stewart$^3$ \email stewarwf@sutterhealth.org
        %\addr Sutter Health
        \AND
        \name Jimeng Sun$^1$ \email jsun@cc.gatech.edu\\
        %\addr Georgia Institute of Technology
        \\$^1$Georgia Institute of Technology \quad $^2$ Vanderbilt University \quad $^3$ Sutter Health}
\maketitle
\vspace{-8mm}

\begin{abstract}
Access to electronic health record (EHR) data has motivated computational advances in medical research. However, various concerns, particularly over privacy, can limit access to and collaborative use of EHR data.  Sharing synthetic EHR data could mitigate risk. %; however, existing generative methods do not capture the rich diversity, subtly or intricacy of associations between medical concepts, which hampers the building of machine learning models with practical utility. 

In this paper, we propose a new approach, medical Generative Adversarial Network (\mname), to generate realistic synthetic patient records. Based on input real patient records, \mname can generate high-dimensional discrete variables (e.g., binary and count features) via a combination of an autoencoder and generative adversarial networks. We also propose minibatch averaging to efficiently avoid mode collapse, and increase the learning efficiency with batch normalization and shortcut connections. To demonstrate feasibility, we showed that \mname generates synthetic patient records that achieve comparable performance to real data on many experiments including distribution statistics, predictive modeling tasks and a medical expert review. We also empirically observe a limited privacy risk in both identity and attribute disclosure using \mname.  
\end{abstract}

\section{Introduction}
\vspace*{-2mm}
\label{sec:intro}
The adoption of electronic health records (EHR) by healthcare organizations (HCOs), along with the large quantity and quality of data now generated, has led to an explosion in \emph{computational health}. 
%In doing so, researchers from a variety fields, including computer science, statistics, and informatics, are partnering with clinical domain experts and industrial affiliates (e.g., pharmaceuticals) to conduct novel discovery-driven research and refine existing healthcare practices. 
%Given the potential, research opportunities are expected to diversify and grow \cite{coorevits2013electronic}. 
However, the wide adoption of EHR systems does not automatically lead to easy access to EHR data for researchers. One
reason behind limited access stems from the fact that EHR data are composed of personal identifiers, which in combination with potentially sensitive medical information, induces privacy concerns. As a result, access to such data for secondary purposes (e.g., research) is  regulated, as well as controlled by the HCOs groups that are at risk if data are misused or breached. The review process by legal departments and institutional review boards can take months, with no guarantee of access \cite{hodge1999legal}. This process limits timely opportunities to use data and may slow advances in biomedical knowledge and patient care \cite{IOMBeyond}.

%There are various ways in which privacy concerns could be addressed, such as asking patients for their consent; however, doing so could bias the population for whom data is available about \cite{elemam09,harris08,hill09,Kho09}. 
%As an alternative, 
HCOs often aim to mitigate privacy risks through the practice of de-identification~\cite{guidance2013}, typically through the perturbation of potentially identifiable attributes (e.g., dates of birth) via generalization, suppression or randomization. \cite{elemam15} 
However, this approach is not impregnable to attacks, such as linkage via residual information to re-identify the individuals to whom the data corresponds~\cite{elemam11b}.
%And, under special circumstances the data resulting from such methods can still be re-identified to the individuals from whom the data was derived \cite{elemam11b}. 
An alternative approach to de-identification is to generate synthetic data \citep{mclachlan2016using, Buczak2010, lombardo2008ta}.
However, realizing this goal in practice has been challenging because the resulting synthetic data are often not sufficiently realistic for machine learning tasks. 
Since many machine learning models for EHR data use aggregated discrete features derived from longitudinal EHRs, we concentrate our effort on generating such aggregated data in this study. Although it is ultimately desirable to generate longitudinal event sequences, in this work we focus on generating high-dimensional discrete variables, which is an important and challenging problem on its own.

\hide{
This work aims at bridging the gap by proposing an efficient algorithm to generate realistic synthetic EHR data. In fact, the raw EHR data consist of clinical event sequences of patients over time. However, for most standard machine learning modeling, those temporal events of a patient are usually aggregated into a static high-dimensional feature vector (in most cases discrete). Although it is ultimately desirable to generate clinical event sequences, this paper focuses on generating high-dimensional discrete feature vectors of patients derived from EHR data.
}
Generative adversarial networks (GANs) have recently been shown to achieve impressive performance in generating high-quality synthetic images \citep{goodfellow2014generative,radford2015unsupervised,goodfellow2016nips}. To understand how, it should first be recognized that a GAN consists of two components: a \textit{generator} that attempts to generate realistic, but fake, data and a \textit{discriminator} that aims to distinguish between the generated fake data and the real data. By playing an adversarial game against each other, the generator can learn the distribution of the real samples - provided that both the generator and the discriminator are sufficiently expressive. Empirically, a GAN outperforms other popular generative models such as variational autoencoders (VAE) \citep{kingma2013auto} and PixelRNN/PixelCNN \citep{van2016pixel,van2016conditional} on the quality of data (i.e., fake compared to real), in this case images, and on processing speed~\cite{goodfellow2016nips}. 
%However, a GAN cannot learn the distribution of discrete variables in its original form. 
However, GANs have not been used for learning the distribution of discrete variables.
 
To address this limitation, we introduce \mname, a neural network model that generates high-dimensional, multi-label discrete variables that represent the events in
%patients'
EHRs (e.g., diagnosis of a certain disease or treatment of a certain medication). Using
%an
EHR source data, \mname is designed to learn the distribution of discrete features, such as diagnosis or medication codes via a combination of an autoencoder and the adversarial framework. %In this setting, the autoencoder is applied to overcome the original GAN's inability to generate discrete samples. 
In this setting, the autoencoder assists the original GAN to learn the distribution of multi-label discrete variables.
The specific contributions of this work are as follows:
\begin{itemize}[noitemsep,topsep=0pt,parsep=0pt,partopsep=0pt]%[leftmargin=5.5mm]
%\vspace*{-3mm}
\item We define
%\mname,
an efficient algorithm to generate high-dimensional multi-label discrete samples by combining an autoencoder with GAN, which we call \mname.
%In particular, \mname
This algorithm is notable in that it handles both binary and count variables.
%\vspace*{-2mm}
%\item \mname translates input EHR data to a program that can generate arbitrarily large volume of high-quality, high-dimensional synthetic patient data. 
%These synthetic data cannot be re-identified, completely mitigating privacy concerns. %\textbf{Brad: I don't think this can be claimed without a proof.  Unfortunately, I haven't had the time to try and work out a proof of this yet.  To make this proof, you really have to state what the adversary has access to (e.g., features of the targeted individuals) and then show that the fake information (and statistical features of such information) do not allow for the recognition of the target individuals.}
%\item Based on real EHR data as input, \mname is able to generate high-quality high-dimensional patient records. Since real patient records data are difficult to access for many researchers, we expect this line of work to have great practical impact.
%\vspace*{-2mm}
\item We propose a simple, yet effective, method called \textit{minibatch averaging} to cope with the situation where GAN overfits to a few training samples (i.e., mode collapse), which outperforms previous methods such as \textit{minibatch discrimination}. %Compared to previous methods%, such as \textit{minibatch discrimination},  our method, \textit{minibatch averaging}, is specifically designed to work with discrete variables and requires no additional parameters.
%\vspace*{-2mm}
\item We demonstrate a close-to-real data performance of \mname using real EHR datasets on a set of diverse tasks, which include reporting distribution statistics, classification performance and medical expert review.
\item We empirically show that \mname leads to acceptable privacy risks in both presence disclosure (i.e., discovery that a patient's record contributed to the GAN) and attribute disclosure (i.e., discovery of a patient's sensitive medical data). %The results indicate that \mname has a promising future in healthcare research.
%, knowing that we are dealing with extremely private, patient-specific records.
\end{itemize}

%In the following section, we discuss past approaches to generate synthetic EHR data and related works on GAN. Then we describe the proposed method \mname and minibatch averaging. In the experiment section we study the performance of \mname via both qualitative and quantitative evaluation. We also address the patient privacy aspect of \mname. We conclude this paper with conclusion and future work. \ecedit{Talk about Pang et al. 2002. Text classification}\jsedit{What is Pang 2002? We should conclude here, if it is a related work, then move to related work section}
\vspace*{-4mm}

\section{Related work}
\vspace*{-2mm}
\label{sec:related}
In this section, we begin with a discussion of existing methods for generating synthetic EHR data.  This is followed by a review recent advances in generative adversarial networks (GANs).  Finally, we summarize specific investigations into generating discrete variables using GANs.

\noindent \textbf{Synthetic Data Generation for Health Data:} De-identification of EHR data is currently the most generally accepted  technical method for protecting patient privacy when sharing EHR data for research in practice \cite{johnson2016mimic}. 
However, de-identification does not guarantee that a system is devoid of risk. In certain circumstances, re-identification of patients can be accomplished through residual distinguishable patterns in various features (e.g., demographics \cite{Sweeney1999,elemam11a}, diagnoses \cite{loukides2010}, lab tests \cite{atreya2013}, visits across healthcare providers \cite{malin2004}, and genomic variants \cite{erlich2014})  To mitigate re-identification vulnerabilities, researchers in the statistical disclosure control community have investigated how to generate synthetic datasets.  Yet, historically, these approaches have been limited to summary statistics for only several variables at a time (e.g., \cite{drechsler,reiter2002}. For instance McLachlan et al.\yrcite{mclachlan2016using} used clinical practice guidelines and health incidence statistics with a state transition machine to generate synthetic patient datasets.
%\textbf{The notion of a synthetic dataset for privacy protection is not new. This has been done in the statistical disclosure community for decades. Look at \url{http://search.proquest.com/openview/0d14a2e049f0bce4ea9f049ee78a0993/1?pq-origsite=gscholar&cbl=105444} }

There is some, but limited, work on synthetic data generation in the healthcare domain and, the majority that has, tend to be disease specific. For example, Buczak et al.~\yrcite{Buczak2010} generated EHRs to explore questions related to the outbreak of specific illnesses, where care patterns in the source EHRs were applied to generate synthetic datasets.
Many of these methods often rely heavily upon domain-specific knowledge along with actual data to generate synthetic EHRs \cite{lombardo2008ta}. %\textbf{Brad: Seems like the transition is off here.} 
More recently, and most related to our work, a privacy-preserving patient data generator was proposed based on a perturbed Gibbs sampler \cite{park2013perturbed}. Still, this approach  can only handle binary variables and its utility was assessed with only a small, low-dimensional dataset. By contrast, our proposed \mname directly captures general EHR data without focusing on a specific disease, which makes it suitable for a greater diversity of applications.

\noindent \textbf{GAN and its Applications:} Attempts to advance GANs \cite{goodfellow2014generative} include, but are not limited to, using convolutional neural networks to improve image processing capacity \cite{radford2015unsupervised}, extending GAN to a conditional architecture for higher quality image generation \cite{mirza2014conditional,denton2015deep,odena2016conditional}, and text-to-image generation \cite{reed2016generative}. %Readers are recommended to refer to \citet{goodfellow2016nips} for a good introduction to GAN.
We, in particular, pay attention to the recent studies that attempted to handle discrete variables using GANs. 

One way to generate discrete variables with GAN is to invoke reinforcement learning. SeqGAN \cite{yu2016seqgan} trains GAN with REINFORCE \cite{williams1992simple} and Monte-Carlo search to generate word sequences. Although REINFORCE enables an unbiased estimation of the gradients of the model via sampling, the estimates come with a high variance. Moreover, SeqGAN focuses on sampling one word (\textit{i.e.} one-hot) at each timestep, whereas our goal is to generate multi-label binary/count variables. Alternatively, one could use specialized distributions, such as the Gumbel-softmax \citep{jang2016categorical, kusner2016gans}, a concrete distribution \citep{maddison2016concrete} or a soft-argmax function \citep{zhanggenerating} to approximate the gradient of the model from discrete samples. However, since these approaches focus on the softmax distribution, they cannot be directly invoked for multi-label discrete variables, especially in the count variable case. 
Yet another way to handle discrete variables is to generate distributed representations, then decode them into discrete outputs. For example, Glover \yrcite{glover2016modeling} generated document embeddings with a GAN, but did not attempt to simulate actual documents. 

To handle high-dimensional multi-label discrete variables, \mname  generates the distributed representations of patient records with a GAN. It then decodes them to simulated patient records with an autoencoder.
\vspace*{-2mm}

\vspace*{-4mm}
\section{Method}
\vspace*{-2mm}
\label{sec:method}
This section begins with a formalization of the structure of EHR data and the corresponding
%formal
mathematical notation we adopt in this work, This is followed by a  detailed description of the \mname algorithm.
\vspace*{-3mm}
\subsection{Description of EHR Data and Notations}
\vspace*{-2mm}
\label{ssec:ehr}
We assume there are $|\mathcal{C}|$ discrete variables (\textit{e.g.}, diagnosis, medication or procedure codes) in the EHR data that can be expressed as a fixed-size vector $\xb \in \mathbb{Z}_{+}^{|\mathcal{C}|}$, where the value of the $i^{th}$ dimension indicates the number of occurrences (\textit{i.e.}, counts) of the $i$-th variable in the patient record. In addition to the count variables, a visit
%record
can also be represented as a binary vector $\xb \in \{0,1\}^{|\mathcal{C}|}$, where the $i^{th}$ dimension indicates the absence or occurrence of the $i^{th}$ variable in the patient record. It should be noted that we can also represent demographic information, such as age and gender, as count and binary variables, respectively.

Learning the distribution of count variables is generally more difficult than learning the distribution of binary variables. This is because the model needs to learn more than simple co-occurrence relations between the various dimensions. Moreover, in EHR data, certain clinical concepts tend to occur much more frequently (e.g., essential hypertension) than others.  This is problematic because it can skew a distribution around different dimensions.

%\textbf{Brad: What do you mean by sequential?  Are you talking about ordered? Temporal? A time series?}
%For the discrete variables, we use binary values instead of count (\textit{i.e.} non-negative integer) variables for three reasons: 1) count values are severely unbalanced\footnote{One patient in our dataset was diagnosed with chronic ulcer of skin 1157 times while other variables were 0} and are heavily affected by the length of the observation, therefore making it difficult for the model to learn the distribution 2) When using aggregate records for prediction tasks, performance is more or less the same when using count variables and binary variables. We provide empirical results in Appendix \ref{appendix:binary} 3) It is much easier for medical experts to conduct qualitative evaluation with binary variables than count variables. \ecedit{Should we talk about sequence generation where all values are binary?} %

\vspace*{-3mm}
\subsection{Preliminary: Generative Adversarial Network}
\vspace*{-2mm}
\label{ssec:gan}
In a GAN, the generator $G(\zb; \theta_g)$ accepts a random prior $\zb \in \mathbb{R}^r$ and generates synthetic samples $G(\zb) \in \mathbb{R}^d$, while the discriminator $D(\xb; \theta_d)$ determines whether a given sample is real or fake. The optimal discriminator $D^*$ would perfectly distinguish real samples from fake samples. The optimal generator $G^*$ would generate fake samples that are indistinguishable from the real samples so that $D$ is forced to make random guesses. Formally, $D$ and $G$ play the following minimax game with the value function $V(G, D)$:
\vspace*{-1mm}
\small
\begin{align*}
\min_{G} \max_{D} V(G,D) & = \mathbb{E}_{\xb \sim p_{data}} [\log D(\xb)] \\ 
& + \mathbb{E}_{\zb \sim p_{\zb}} [\log (1 - D(G(\zb)))] \label{eq:gan}
%\vspace*{-1mm}
\end{align*}
\normalsize
where $p_{data}$ is the distribution of the real samples and $p_{\zb}$ is the distribution of the random prior, for which $\mathcal{N}(0,1)$ is generally used. Both $G$ and $D$ iterate in optimizing the respective parameters $\theta_g$ and $\theta_d$ as follows,
\vspace*{-2mm}
\small
\begin{align}
\theta_d & \leftarrow \theta_d + \alpha \nabla_{\theta_d}\frac{1}{m}\sum_{i=1}^{m} \log D(\xb_i) + \log (1 - D(G(\zb_i))) \nonumber \\
\theta_g & \leftarrow \theta_g - \alpha \nabla_{\theta_g}\frac{1}{m}\sum_{i=1}^{m} \log (1 - D(G(\zb_i))) \nonumber
\vspace*{-8mm}
\end{align}
\normalsize
where $m$ is the size of the minibatch and $\alpha$ the step size. In practice, however, $G$ can be trained to maximize $\log(D(G(\zb))$ instead of minimizing $\log(1 - D(G(\zb))$ to provide stronger gradients in the early stage of the training \citep{goodfellow2014generative} as follows,
\vspace*{-3mm}
\small
\begin{equation}
\theta_g \leftarrow \theta_g + \alpha \nabla_{\theta_g}\frac{1}{m}\sum_{i=1}^{m} \log D(G(\zb_i)) \label{eq:alternative_G}
\vspace*{-2mm}
\end{equation}
\normalsize
Henceforth, we use Eq.\eqref{eq:alternative_G} as it showed significantly more stable performance in our investigation. We also assume throughout the paper that both $D$ and $G$ are implemented with feedforward neural networks.

\begin{SCfigure}
\centering
\includegraphics[width=0.4\textwidth]{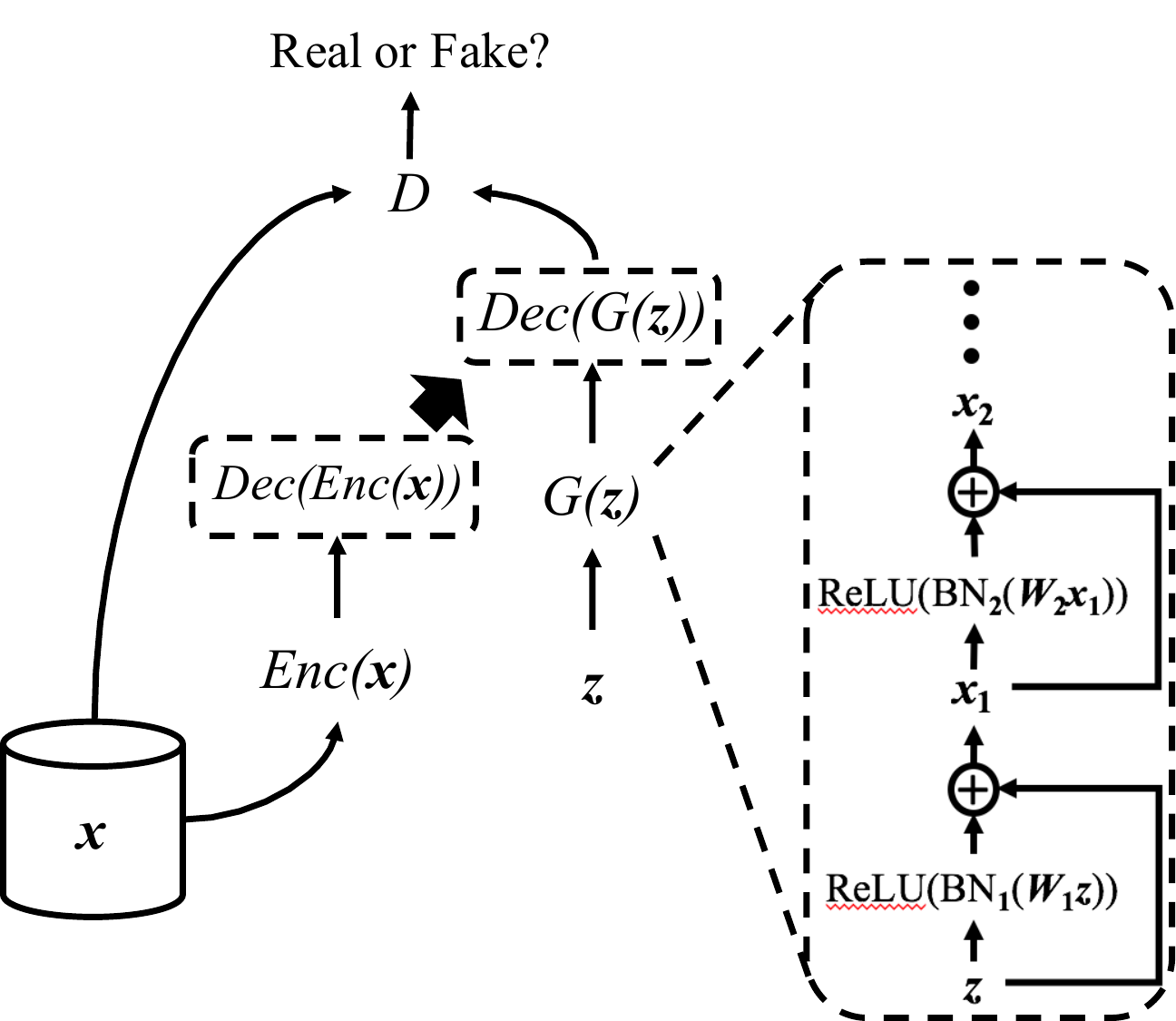}
\vspace*{-3mm}
\caption{Architecture of \mname: The discrete $\xb$ comes from the source EHR data, $\zb$ is the random prior for the generator $G$; $G$ is a feedforward network with shortcut connections (right-hand side figure); An autoencoder (i.e, the encoder $Enc$ and decoder $Dec$) is learned from $\xb$; The same decoder $Dec$ is used after the generator $G$ to construct the discrete output. The discriminator $D$ tries to differentiate real input $\xb$ and discrete synthetic output $Dec(G(\zb))$.}
\vspace*{-2mm}
\label{fig:medgan}
\end{SCfigure}
\vspace*{-2mm}
\subsection{\mname}
\label{ssec:medgan}
\vspace*{-2mm}
Since the generator $G$ is trained by the error signal from the discriminator $D$ via backpropagation, the original GAN can only learn to approximate discrete patient records $\xb \in \mathbb{Z}_{+}^{|\mathcal{C}|}$ with continuous values.
%the original GAN cannot directly learn the distribution of discrete patient records $\xb \in \mathbb{Z}_{+}^{|\mathcal{C}|}$, \ecedit{but can only learn to approximate discrete values with continuous values.}
We alleviate this limitation by leveraging the autoencoder. Autoencoders are trained to project given samples to a lower dimensional space, then project them back to the original space.
Such a mechanism leads the autoencoder to learn salient features of the samples and has been successfully used in certain applications, such as image processing \cite{Goodfellow-et-al-2016,vincent2008extracting}. 

In this work, We apply the autoencoder to learn the salient features of discrete variables that can be applied to decode the continuous output of $G$. This allows the gradient flow from $D$ to the decoder $Dec$ to enable the end-to-end fine-tuning. As depicted by Figure \ref{fig:medgan}, an autoencoder consists of an encoder $Enc(\xb; \theta_{enc})$ that compresses the input $\xb \in \mathbb{Z}_{+}^{|\mathcal{C}|}$ to $Enc(\xb) \in \mathbb{R}^h$, and a decoder $Dec(Enc(\xb); \theta_{dec})$ that decompresses $Enc(\xb)$ to $Dec(Enc(\xb))$ as the reconstruction of the original input $\xb$. The objective of the autoencoder is to minimize the reconstruction error:
\vspace*{-3mm}
\small
\begin{align}
& \frac{1}{m} \sum_{i=0}^{m} ||\xb_i - \xb'_i||^{2}_{2} \label{eq:ae_count} \\
& \frac{1}{m} \sum_{i=0}^{m} \xb_i \log \xb'_i + (1 - \xb_i) \log ( 1 - \xb'_i) \label{eq:ae_binary} \\
& \mbox{where } \xb'_i = Dec(Enc(\xb_i)) \nonumber
\vspace*{-3mm}
\end{align}
\normalsize
where $m$ is the size of the mini-batch. We use the mean squared loss (Eq.\eqref{eq:ae_count}) for count variables and cross entropy loss (Eq.\eqref{eq:ae_binary}) for binary variables. For count variables, we use rectified linear units (ReLU) as the activation function in both $Enc$ and $Dec$. For binary variables, we use tanh activation for $Enc$ and the sigmoid activation for $Dec$.\footnote{We considered a denoising autoencoder (dAE) \citep{vincent2008extracting} as well, but there was no discernible improvement in performance.}% due to the sparse nature of the EHR data. 

With the pre-trained autoencoder, we can allow GAN to generate distributed representation of patient records (i.e., the output of the encoder $Enc$), rather than generating patient records directly. Then the pre-trained decoder $Dec$ can pick up the right signals from $G(\zb)$ to convert it to the patient record $Dec(G(\zb))$.
%As the generator $G$ and the encoder $Enc$ both generate continuous values, the decoder $Dec$ can pick up the right signals to convert synthetic continuous samples $G(\zb) \in \mathbb{R}^h$ to  samples $Dec(G(\zb)) \in \mathbb{Z}_{+}^{|\mathcal{C}|}$. 
The discriminator $D$ is trained to determine whether the given input is a synthetic sample $Dec(G(\zb))$ or a real sample $\xb$. The architecture of the proposed model \mname is depicted in Figure \ref{fig:medgan}. \mname is trained in a similar fashion as the original GAN as follows,
\vspace*{-2mm}
\small
\begin{align}
& \theta_d \leftarrow \theta_d + \alpha \nabla_{\theta_d}\frac{1}{m}\sum_{i=1}^{m} \log D(\xb_i) + \log (1 - D(\xb_{\zb_i})) \nonumber \\
& \theta_{g, dec} \leftarrow \theta_{g, dec}+ \alpha \nabla_{\theta_{g, dec}}\frac{1}{m}\sum_{i=1}^{m} \log D(\xb_{\zb_i}) \nonumber \\
& \mbox{where } \xb_{\zb_i} = Dec(G(\zb_i)) \nonumber
\vspace*{-2mm}
\end{align}
\normalsize
It should be note that we can round the values of $Dec(G(\zb))$ to their nearest integers to ensure that the discriminator $D$ is trained on discrete values instead of continuous values. We experimented both with and without rounding and empirically found that training $D$ in the latter scenario
%without explicit rounding
led to better predictive performance in section \ref{ssec:quant_results_binary}. Therefore, we assume, for the remainder of this paper, that $D$ is trained without explicit rounding.

We fine-tune the pre-trained parameters of the decoder $\theta_{dec}$ while optimizing for $G$. Therefore, the generator $G$ can be viewed as a neural network with an extra hidden layer pre-trained to map continuous samples to discrete samples. We used ReLU for all of $G$'s activation functions, except for the output layer, where we used the tanh function\footnote{We also applied tanh activation for the encoder $Enc$ for consistency.}. For $D$, we used ReLU for all activation functions except for the output layer, where we used the sigmoid function for binary classification. 
\vspace*{-3mm}

\subsection{Minibatch Averaging}
\vspace*{-2mm}
\label{ssec:minibatch_averaging}
%\jsedit{I suggest we expand this section with a picture and possibly more illustration as this is the main novelty of this paper. We should also remove 3.2 GAN (merge the necessary content into the later section)}
Since the objective of the generator $G$ is to produce samples that can fool the discriminator $D$, $G$ could learn to map different random priors $\zb$ to the same synthetic output, rather than producing diverse synthetic outputs. This problem is denoted as \textit{mode collapse}, which arises most likely due to the GAN's optimization strategy often solving the max-min problem instead of the min-max problem \cite{goodfellow2016nips}. Some methods have been proposed to cope with mode collapse (e.g., minibatch discrimination and unrolled GANs), but they require
%some
\emph{ad hoc} fine-tuning of the hyperparameters and scalability is often neglected \citep{salimans2016improved,metz2016unrolled}. 

By contrast, \mname offers a simple and efficient method to cope with mode collapse when generating discrete outputs. Our method, \textit{minibatch averaging}, is motivated by the philosophy behind minibatch discrimination. It allows the discriminator $D$ to view the minibatch of real samples $\xb_1, \xb_2, \ldots$ and the minibatch of the fake samples $G(\zb_1), G(\zb_2), \ldots$, respectively, while classifying a real sample and a fake sample. Given a sample to discriminate, minibatch discrimination calculates the distance between the given sample and every sample in the minibatch in the latent space. Minibatch averaging, by contrast, provides the average of the minibatch samples to $D$, modifying the objective as follows:
\vspace*{-2mm}
\small
\begin{align}
& \theta_d \leftarrow \theta_d + \alpha \nabla_{\theta_d}\frac{1}{m}\sum_{i=1}^{m} \log D(\xb_i, \bar{\xb}) + \log (1 - D(\xb_{\zb_i}, \bar{\xb}_{\zb})) \nonumber \\
& \theta_{g, dec} \leftarrow \theta_{g, dec}+ \alpha \nabla_{\theta_{g, dec}}\frac{1}{m}\sum_{i=1}^{m} \log D(\xb_{\zb_i}, \bar{\xb}_{\zb}) \nonumber \\
& \mbox{where } \bar{\xb} = \frac{1}{m} \sum_{i=1}^{m}\xb_i \mbox{,\quad} \xb_{\zb_i} = Dec(G(\zb_i)) \mbox{,\quad} \bar{\xb}_{\zb} = \frac{1}{m} \sum_{i=1}^{m} \xb_{\zb_i}\nonumber
\vspace*{-2mm}
\end{align}
\normalsize
where $m$ denotes the size of the minibatch. Specifically, the average of the minibatch is concatenated on the sample and provided to the discriminator $D$. 

\noindent{\bf Binary variables:} When processing binary variables $\xb \in \{0,1\}^{|\mathcal{C}|}$, the average of minibatch samples $\bar{\xb}$ and $\bar{\xb}_{\zb}$ are equivalent to the maximum likelihood estimate of the Bernoulli success probability $\hat{p}_k$ of each dimension $k$.
This information makes it easier for $D$ to ascertain whether a given sample is real or fake, if $\hat{p}_k$'s of fake samples are considerably different from those of real samples. This is especially likely when mode collapse occurs because the $\hat{p}_k$'s for most dimensions of the fake samples become dichotomized (either 0 or 1), whereas the $\hat{p}_k$'s of real samples generally take on a value between 0 and 1. Therefore, if $G$ wants to fool $D$, it will have to generate more diverse examples within the minibatch $Dec(G(\zb_1, \zb_2, \ldots))$. 

\noindent{\bf Count variables:} Count variables are a more accurate description of clinical events. They can indicate the number of times a certain diagnosis was made or a certain medication was prescribed over multiple hospital visits. For count variables $\xb \in \mathbb{Z}_{+}^{|\mathcal{C}|}$, the average of minibatch samples $\bar{\xb}$ and $\bar{\xb}_{\zb}$ can be viewed as the estimate of the binomial distribution mean $n\widehat{p_k}$ of each dimension $k$, where $n$ is the number of hospital visits. Hence minibatch averaging for the count variables also provides helpful statistics to the discriminator $D$, %\footnote{Although, the estimated probability $\hat{p}_k$ in the binary variable case is \textit{per patient}, but is \textit{per visit} in the count variable case.}, 
guiding the generator $G$ to generate more diverse and realistic samples. As our experiments show, minibatch averaging works surprisingly well and does not require additional parameters like minibatch discrimination. As a consequence, it has minimal impact to the training time. It is further worth mentioning that, for both binary and count variables, a minibatch that is larger than usual is recommended to properly capture the statistics of the real data. We use 1,000 records for a minibatch in this investigation.

\vspace*{-2mm}
\subsection{Enhanced Generator Training}
\vspace*{-2mm}
%Goodfellow \yrcite{goodfellow2016nips} states that, because the optimality of the discriminator $D$ could be the necessary condition to train GAN, it is fine for $D$ to overpower the generator $G$. However, many believe that the balance between $D$'s power and $G$'s power should be somehow maintained in order for $G$ to effectively learn the distribution of the real samples \cite{goodfellow2016nips}. One can choose different model sizes for $D$ and $G$ to control the power balance. In image processing, $D$ is usually deeper than $G$ and has more filter per layer in order to guarantee some accuracy for $D$.
Similar to image processing GANs, we observed that balancing the power of $D$ and $G$ in the multi-label discrete variable setting was quite challenging \cite{goodfellow2016nips}. Empirically, we observed that training \mname with minibatch averaging demonstrated $D$ consistently overpowering $G$ after several iterations. While $G$ still managed to learn under such situation, the performance seemed suboptimal, and updating $\theta_g$ and $\theta_{dec}$ more often than $\theta_d$ in each iteration only degraded performance. Considering the importance of an optimal $D$ \cite{goodfellow2016nips}, we chose not to limit the discriminative power of $D$, but rather improve the learning efficiency of $G$ by applying batch normalization \cite{ioffe2015batch} and shortcut connection \cite{he2016deep}. $G$'s $k^{th}$ layer is now formulated as follows:
\vspace{-2mm}
\begin{equation*}
\xb_k = \mbox{ReLU}(\mbox{BN$_{k}$}(\Wb_k \xb_{k-1})) + \xb_{k-1} \label{eq:batch_norm}
\vspace{-2mm}
\end{equation*}
where ReLU is the rectified linear unit, BN$_{k}$ is the batch normalization at the $k$-th layer, $\Wb_k$ is the weight matrix of the $k$-th layer, and $\xb_{k-1}$ is the input from the previous layer. The right-hand side of Figure \ref{fig:medgan} depicts the first two layers of $G$. Note that we do not incorporate the bias variable into each layer because batch normalization negates the necessity of the bias term. Additionally, batch normalization and shortcut connections could be applied to the discriminator $D$, but the experiments showed that $D$ was consistently overpowering $G$ without such techniques, and we empirically found that a simple feedforward network was sufficient for $D$. We describe the overall optimization algorithm in the Appendix~\ref{appendix:algorithm}.

\vspace*{-3mm}
\subsection{Privacy Consideration}
\vspace*{-2mm}
\label{sec:privacy}
%Traditionally, real patient records are released with certain modification such as data perturbation and generalization. The key difference to \mname is that the released dataset in the traditional setting still has a clear mapping to the original patients (such a mapping is often one to one). However, the mapping between the generated data from \mname and the training data of specific patients is not explicit. Intuitively, it seems to imply the privacy of the training patients can be better preserved with \mname.
When EHRs are de-identfied via methods such generalization or randomization, there often remains a 1-to-1 mapping to the underlying records from where they were derived.
%And to functionally achieve de-identification,  masking of the data is applied, such as generalization or randomization to the original values.  
%perturbation or generalization, but the released dataset often has a clear mapping (often 1-to-1) to the original patients. 
However, in our case, the mapping between the generated data from \mname and the training data of specific patients is not explicit. Intuitively, this seems to imply that the privacy of the patients can be better preserved with \mname; however,
it also begs the question of how to evaluate the privacy in the system. We perform a formal assessment of \mname's privacy risks based on two definitions of privacy.
%There are two aspects of privacy risk involved in deploying a data generator like \mname: 

\noindent \textbf{Presence disclosure} occurs when an attacker can determine that \mname was trained with a dataset including the record from patient $x$. \cite{nergiz:10} Presence disclosure for \mname happens when a powerful attacker, one who already possesses the complete records of a set of patients $P$, can determine whether anyone from $P$ are in the training set by observing the generated patient records. More recently, for machine learned models, this has been referred to as an \emph{membership inference attack} \cite{Shokri:17}. 
%Even before empirical evaluation, we can already argue that this strong assumption (e.g. attacker's complete knowledge) makes such an attack plausible, but
the knowledge gained by the attacker may be limited, if the dataset is well balanced in its clinical concepts.
%is  limited, disclosure may occur when this attack is based on features that are devoid of medical. 
\noindent \textbf{Attribute disclosure} occurs when attackers can derive additional attributes such as diagnoses and medications about patient $x$ based on a subset of attributes they already know  about $x$. \cite{Matwin2015} We believe that attribute disclosure for \mname could be a more prominent issue because the attacker only needs to know a subset of attributes of a patient. Moreover, the goal of the attacker is to gain knowledge of the unknown attributes by observing similar patients generated by \mname.

Considering the difficulty of deriving analytic proof of privacy for GANs and simulated data, we report the empirical analysis of both risks to understand the extent to which privacy can be achieved, as commonly practiced in the statistical disclosure control community. \cite{Domingo-Ferrer2003}
\vspace*{-5mm}

\section{Experiments}
\vspace*{-2mm}
\label{sec:exp}
We evaluated \mname with three distinct EHR datasets. First, we describe the datasets and baseline models. Next, we report the quantitative evaluation results using both binary and count variables. We then perform a qualitative analysis through medical expert review. Finally, we address the privacy aspect of \mname. The source code of \mname is publicly available at \url{https://github.com/mp2893/medgan}.
\vspace*{-3mm}
\subsection{Experimental Setup}
\label{ssec:exp_setup}
\vspace*{-2mm}
\begin{table}[t]
% \small
\vspace*{-3mm}
\caption{Basic statistics of datasets A, B and C}
\vspace*{-2mm}
\centering
\begin{footnotesize}
\begin{tabular}{l|c|c|c}
\toprule
\textbf{Dataset} & \textbf{(A) Sutter PAMF} & \textbf{(B) MIMIC-III} & \textbf{(C) Sutter Heart Failure}\\
\toprule
\# of patients & 258,559 & 46,520 & 30,738 \\ 
\# of unique codes & 615 & 1071 & 569\\
Avg. \# of codes per patient & 38.37 & 11.27 & 53.02\\
Max \# of codes for a patient & 198 & 90 & 871\\
Min \# of codes for a patient & 1 & 1 & 2\\
\bottomrule
\end{tabular}
\end{footnotesize}
\vspace*{-4mm}
\label{tab:data_stats}
\end{table}
\textbf{Source data: } The datasets in this study were from A) Sutter Palo Alto Medical Foundation (PAMF), which consists of 10-years of longitudinal medical records of 258K patients, B) the MIMIC-III dataset \citep{johnson2016mimic, goldberger2000physiobank}, which is a publicly available dataset consisting of the medical records of 46K intensive care unit (ICU) patients over 11 years old and C) a heart failure study dataset from Sutter, which consists of 18-months observation period of 30K patients. From dataset A and C, we extracted diagnoses, medications and procedure codes, which were then respectively grouped by Clinical Classifications Software (CCS) for ICD-9\footnote{https://www.hcup-us.ahrq.gov/toolssoftware/ccs/ccs.jsp}, Generic Product Identifier Drug Group\footnote{http://www.wolterskluwercdi.com/drug-data/medi-span-electronic-drug-file/} and
%Clinical Classifications Software
for CPT\footnote{https://www.hcup-us.ahrq.gov/toolssoftware/ccs\_svcsproc/ ccssvcproc.jsp}. From dataset B, we extracted ICD9 codes only and grouped them by generalizing up to their first 3 digits. Finally, we aggregate a patient's longitudinal record into a single fixed-size vector $\xb \in \mathbb{Z}_{+}^{|\mathcal{C}|}$, where $|\mathcal{C}|$ equals 615, 1071 and 569 for dataset A, B and C respectively. Note that datasets A and B are binarized for experiments regarding binary variables while dataset C is used for experiments regarding count variables. A summary of the datasets are in Table \ref{tab:data_stats}.

\noindent \textbf{Models for comparison: } To assess the effectiveness of our methods, we tested multiple versions of \mname:
%\vspace*{-4mm}
\begin{itemize}[noitemsep,topsep=0pt,parsep=0pt,partopsep=0pt]
\item %\textbf{GAN:} 
\textbf{GAN:} We use the same architecture as \mname with the standard training strategy, but do not pre-train the autoencoder. %This is equivalent to the original GAN where we use multi-layer perceptrons for the generator $G$.
%\vspace*{-2mm}
\item \textbf{GAN$_P$:} We pre-train the autoencoder (in addition to the GAN).
%\vspace*{-2mm}
\item \textbf{GAN$_{PD}$:} We pre-train the autoencoder and use minibatch discrimination \citep{salimans2016improved}.
%\vspace*{-2mm}
\item \textbf{GAN$_{PA}$:} We pre-train the autoencoder and use minibatch averaging.
%\vspace*{-2mm}
\item \textbf{\mname:} We pre-train the autoencoder and use minibatch averaging. We also use batch normalization and a shortcut connection for the generator $G$.
%\vspace*{-2mm}
\end{itemize}
We also compare the performance of \mname with several popular generative methods as below.
%\vspace*{-2mm}
\begin{itemize}[noitemsep,topsep=0pt,parsep=0pt,partopsep=0pt]
\item \textbf{Random Noise (RN):} Given a real patient record $\xb$, we invert the binary value of each code (i.e., dimension) with probability 0.1. This is not strictly a generative method, but rather it is a simple implementation of a privacy protection method based on randomization. %However, we include this approach as it is a common practice in the medical domain. \textbf{Brad: Not really.  I don't know of any methods that actually do this... I wouldn't say that it's common practice.  You might say that its a naive implementation of a privacy protection method based on randomization.}
\item \textbf{Independent Sampling (IS):} For the binary variable case, we calculate the Bernoulli success probability of each code in the real dataset, based on which we sample binary values to generate the synthetic dataset. For the count variable case, we use the kernel density estimator (KDE) for each code then sample from that distribution.
\item \textbf{Stacked RBM (DBM):} We train a stacked Restricted Boltzmann Machines \citep{hinton2006reducing}, then, using Gibbs sampling, we can generate synthetic binary samples. 
There are studies that extend RBMs beyond binary variables \cite{hinton2009replicated,gehler2006rate,tran2011mixed}. In this work, however, as our goal is to study \mname's performance in various aspects, we use the original RBM only.
\item \textbf{Variational Autoencoder (VAE):} We train a variational autoencoder \cite{kingma2013auto} where the encoder and the decoder are constructed with feed-forward neural networks.
\end{itemize}

\noindent \textbf{Implementation details:} We implemented \mname with TensorFlow 0.12 \citep{tensorflow2015-whitepaper}. For training models, we used Adam \citep{kingma2014adam} with the learning rate set to 0.001, and a mini-batch of 1,000 patients on a machine equipped with Intel Xeon E5-2630, 256GB RAM, four Nvidia Pascal Titan X's and CUDA 8.0. The hyperparameter details are provided in Appendix \ref{appendix:hyperparameter}.
\vspace*{-3mm}

\subsection{Quantitative Evaluation for Binary Variables}
\label{ssec:quant_results_binary}
\vspace*{-2mm}
We evaluate the model performance for binary variables in this section, and provide the evaluation results of count variables in Appendix \ref{appendix:exp_quant_count}.
For all evaluations, we divide the dataset into a training set $R \in \{0,1\}^{N \times |\mathcal{C}|}$ and a test set $T \in \{0,1\}^{n \times |\mathcal{C}|}$ by 4:1 ratio. We use $R$ to train the models, then generate synthetic samples $S \in \{0,1\}^{N \times |\mathcal{C}|}$ that are assessed in various tasks. For \mname and VAE, we round the values of the generated dataset to the nearest integer values.
\begin{itemize}[leftmargin=5.5mm]
\vspace*{-2mm}
\item \textbf{Dimension-wise probability: } This is a basic sanity check to confirm the model has learned each dimension's distribution correctly. We use the training set $R$ to train the models, then generate the same number of synthetic samples $S$. Using $R$ and $S$, we compare the Bernoulli success probability $p_k$ of each dimension $k$. 
\vspace*{-2mm}
\item \textbf{Dimension-wise prediction: } This task indirectly measures how well the model captures the inter-dimensional relationships of the real samples. After training the models with $R$ to generate $S$, we choose one dimension $k$ to be the label $\yb_{R_k} \in \{0,1\}^{N}$ and $\yb_{S_k} \in \{0,1\}^{N}$. The remaining $R_{\backslash k} \in \{0,1\}^{N \times |\mathcal{C}|-1}$ and $S_{\backslash k} \in \{0,1\}^{N \times |\mathcal{C}|-1}$ are used as features to train two logistic regression classifiers LR$_{R_k}$ and LR$_{S_k}$ to predict $\yb_{R_k}$ and $\yb_{S_k}$,  respectively. Then, we use the model LR$_{R_k}$ and LR$_{S_k}$ to predict label $\yb_{T_k} \in \{0,1\}^{n}$ of the test set $T$.
% given the remaining $T_{\backslash k} \in \{0,1\}^{n \times |\mathcal{C}|-1}$. 
We can assume that the closer the performance of LR$_{S_k}$ to that of LR$_{R_k}$, the better the quality of the synthetic dataset $S$. 
%This evaluation is important because in a real-world application, researchers are likely to use only the synthetic data due to legal concerns. And the synthetic data will have more value if the learned knowledge from it could be transferred to the applications that use the real data. 
We use F1-score to measure the prediction performance, with the threshold set to 0.5. 
%We also provide the area under the precision-recall curve (AUCPR) score in the Appendix \ref{appendix:performance}.
\vspace*{-2mm}
\end{itemize}
To mitigate the repetition of results, we present our evaluation of dataset A in this section and direct the reader to Appendix \ref{appendix:performance} for the results from dataset B.

\begin{figure}[t]
\centering
\begin{subfigure}{.8\textwidth}
  \centering\captionsetup{width=1\linewidth}%
  \includegraphics[width=1\linewidth]{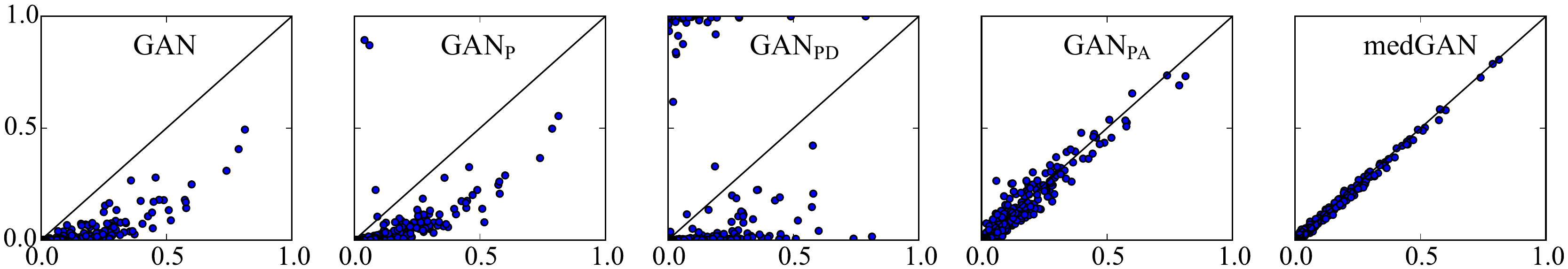}
  \caption{Dimension-wise probability performance of various versions of \mname.}
  \label{fig:sutter_dimprob_gans}
\end{subfigure}
\begin{subfigure}{.8\textwidth}
  \centering\captionsetup{width=1\linewidth}%
  \includegraphics[width=1\linewidth]{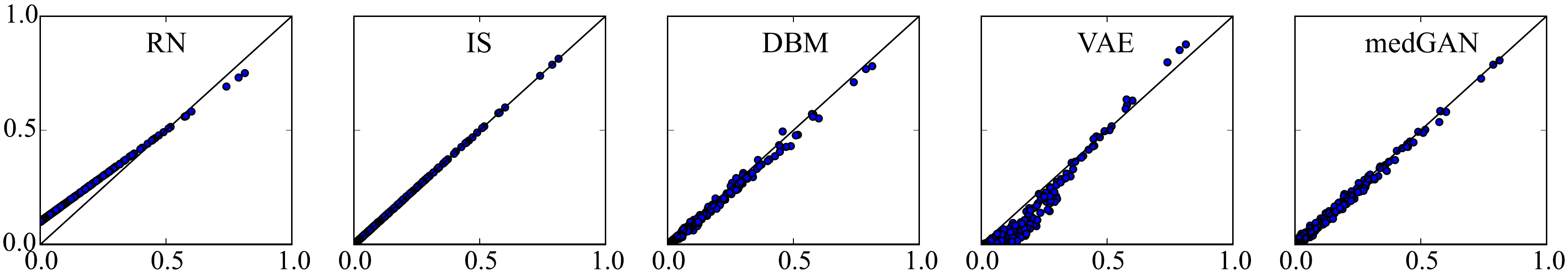}
  \caption{Dimension-wise probability performance of baseline models and \mname.}
  \label{fig:sutter_dimprob_baselines}
\end{subfigure}
\vspace*{-2mm}
\caption{Scatterplots of dimension-wise probability results. Each dot represents one of 615 codes. The x-axis represents the Bernoulli success probability for the real dataset A, and y-axis the probability for the synthetic counterpart generated by each model. The diagonal line indicates the ideal performance where the real and synthetic data show identical quality.}
\vspace*{-5mm}
\label{fig:sutter_binary_dimprob}
\end{figure}
\vspace*{-2mm}
\subsubsection{Dimensions-wise probability}
\vspace*{-2mm}
There are several notable findings that are worth highlighting. The dimension-wise probability performance increased as we used more advanced versions of \mname, where the full \mname shows the best performance as depicted by figure \ref{fig:sutter_dimprob_gans}.
Note that minibatch averaging significantly increases the performance. Since minibatch averaging provides Bernoulli success probability information of real data to the model during training, it is natural that the generator learns to output synthetic data that follow a similar distribution. Minibatch discrimination does not seem to improve the results. This is most likely due to the discrete nature of the datasets. Improving the learning efficiency of the generator $G$ with batch normalization and shortcut connection clearly helped improve the results.

Figure \ref{fig:sutter_dimprob_baselines} compares the dimension-wise probability performance of baseline models with \mname. Independent sampling (IS) naturally shows great performance as expected. DBM, given its stochastic binary nature, shows comparable performance as \mname. VAE, although slightly inferior to DBM and \mname, seems to capture the dimension-wise distribution relatively well, showing specific weakness at processing codes with low probability. Overall, we can see that \mname clearly captures the independent distribution of each code.

\begin{figure}[t]
\centering
\begin{subfigure}{.8\textwidth}
  \centering\captionsetup{width=1\linewidth}%
  \includegraphics[width=1\linewidth]{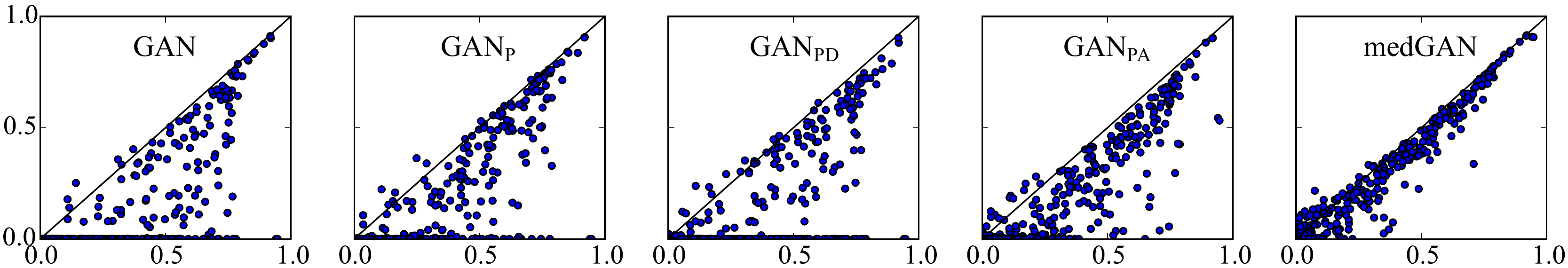}
  \caption{Dimension-wise prediction performance of various versions of \mname.}
  \label{fig:sutter_f1_gans}
\end{subfigure}
\begin{subfigure}{.8\textwidth}
  \centering\captionsetup{width=1\linewidth}%
  \includegraphics[width=1\linewidth]{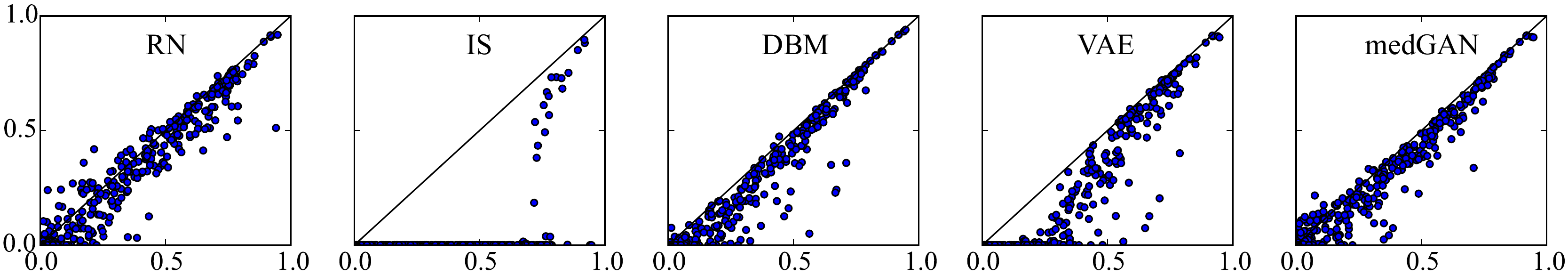}
  \caption{Dimension-wise prediction performance of baseline models and \mname.}
  \label{fig:sutter_f1_baselines}
\end{subfigure}
\vspace*{-2mm}
\caption{Scatterplots of dimension-wise prediction results. Each dot represents one of 615 codes. The x-axis represents the F1-score of the logistic regression classifier trained on the real dataset A. The y-axis represents the F1-score of the classifier trained on the synthetic counterpart generated by each model. The diagonal line indicates the ideal performance where the real and synthetic data show identical quality.}
\vspace*{-5mm}
\label{fig:sutter_binary_f1}
\end{figure}
\vspace*{-4mm}
\subsubsection{Dimensions-wise prediction}
\vspace*{-3mm}
\label{sssec:dimwise_prediction}
Figure \ref{fig:sutter_f1_gans} shows the dimension-wise prediction performance of various versions of \mname. The full \mname again shows the best performance as it did in the dimension-wise probability task. Although the advanced versions of \mname do not seem to dramatically increase the performance as they did for the previous task, this is due to the complex nature of inter-dimensional relationship compared to the independent dimension-wise probability.
Figure \ref{fig:sutter_f1_baselines} shows the dimension-wise prediction performance of baseline models compared to \mname. As expected, IS is incapable of capturing the inter-dimensional relationship, given its naive sampling method. VAE shows similar behavior as it did in the previous task, showing weakness at predicting codes with low occurrence probability. Again, DBM shows comparable, if not slightly better performance to \mname, which seems to come from its binary nature. 
\vspace*{-2mm}

% \subsection{Quantitative Evaluation for Count Variables}
% \label{ssec:quant_results_count}
% \vspace*{-2mm}
% \input{exp_quant_count.tex}

\vspace*{-2mm}
\subsection{Qualitative Evaluation for Count Variables}
\vspace*{-2mm}
\label{ssec:qual_results}
\begin{wrapfigure}{L}{0.25\textwidth}
\centering
\includegraphics[width=0.2\textwidth]{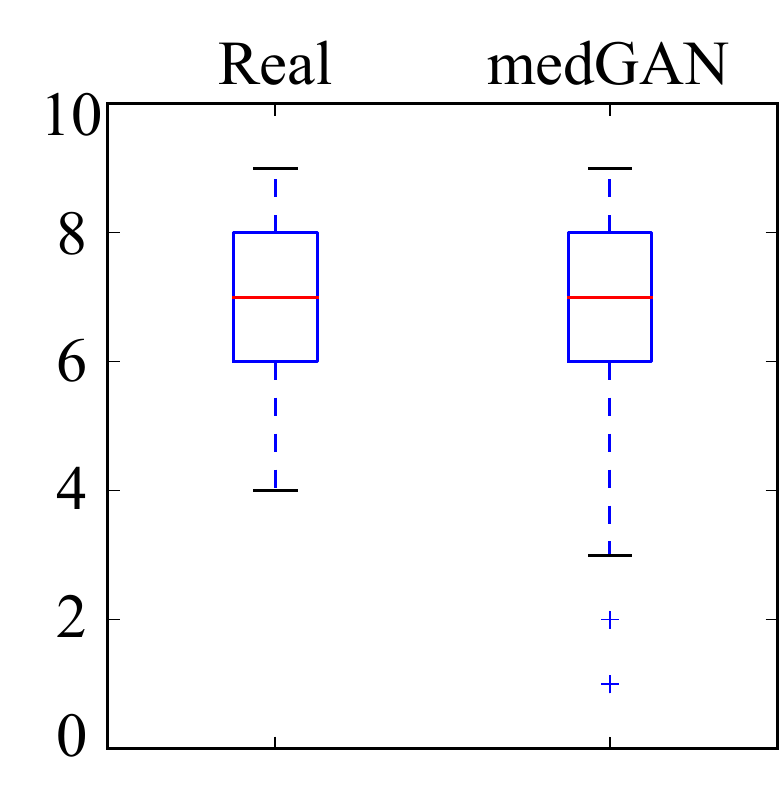}
\vspace*{-3mm}
\caption{Boxplot of the impression scores from a medical expert.} %The synthetic data from \mname provides comparable impression scores in terms of realisticness level.}
\vspace*{-4mm}
\label{fig:boxplot}
\end{wrapfigure}

We conducted a qualitative evaluation of \mname with the help from a medical doctor. A discussion with the doctor taught us that count data are easier to assess its \textit{realistic-ness} than binary data. 
Therefore we use dataset C to train \mname and generate synthetic count samples.
%Therefore we use a subset of PAMF\footnote{We provide the details of how we selected the subset in Appendix \ref{appendix:hf_cohort}.} to train \mname and generate synthetic count samples.
%To conduct a qualitative evaluation of \mname, we use a subset of PAMF to train \mname and generate synthetic samples with count variables. 
In this experiment, we randomly pick 50 records from real data and 50 records from synthetic data, randomly shuffle the order, present them to a medical doctor (specialized in internal medicine) who is asked to score how realistic each record is using scale 1 to 10 (10 being most realistic). %The doctor is allowed to first see another 10 real records to learn their pattern before scoring.% \textbf{Brad: Doctor???}
Here the human doctor is served as the role of discriminator to provide the quality assessment of the synthetic data generated by \mname.

The results of this assessment is shown in Figure \ref{fig:boxplot}. The findings suggest that \mname's synthetic data are generally indistinguishable to a human doctor except for several outliers.
%The real records scored 6.98 on average, while the synthetic records scored 6.36 on average. 
In those cases, the fake records identified by the doctor either lacked appropriate medication codes, or had both male-related codes (\textit{e.g.} prostate cancer) and female-related codes (\textit{e.g.} menopausal disorders) in the same record. The former issue also existed in some of the real records due to missing data, but the latter issue demonstrates a current limitation in \mname which could potentially be alleviated by domain specific heuristics. 
In addition to \mname's impressive performance in statistical aspects, this medical review lends credibility to the qualitative aspect of \mname.

\vspace*{-2mm}
\subsection{Privacy Risk Evaluation}
\vspace*{-2mm}
\label{ssec:privacy_evaluation}
\begin{figure}[ht]
\centering
\includegraphics[width=0.8\textwidth]{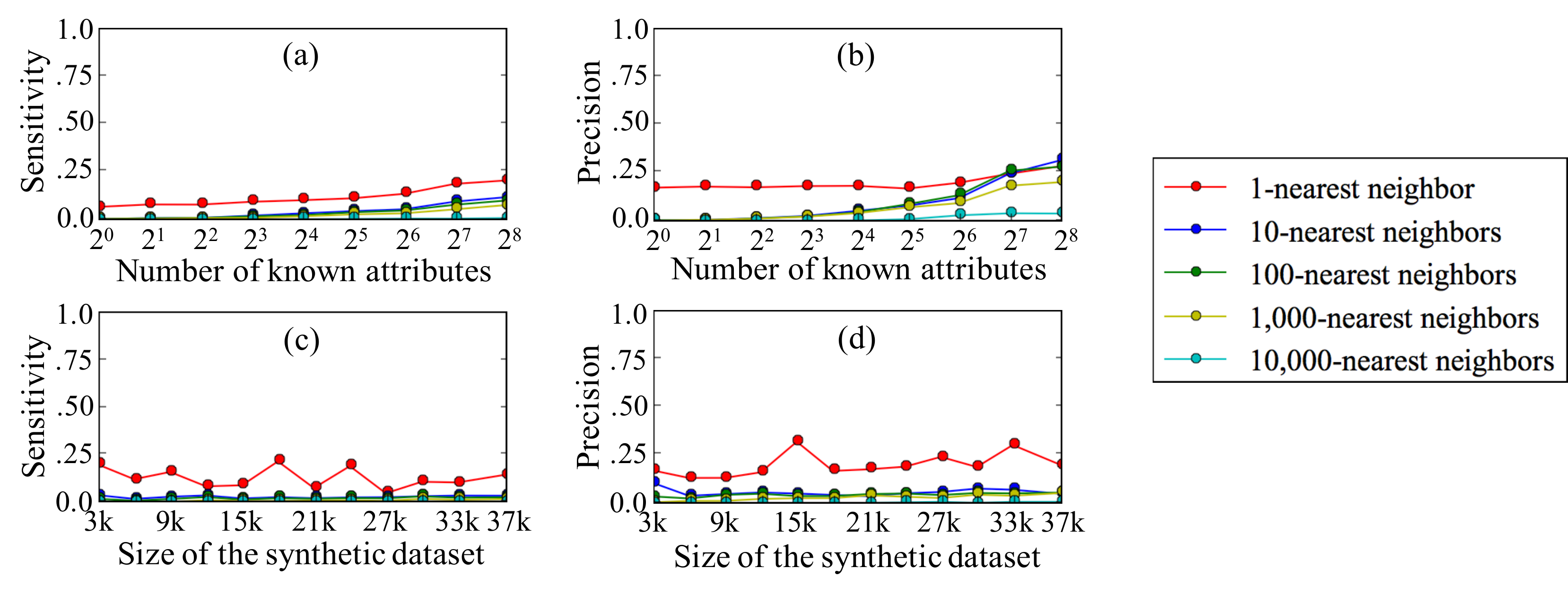}
\vspace*{-2mm}
\caption{\textbf{a,b:} Sensitivity and precision when varying the number of known attributes. The total number of attributes (\textit{i.e.} codes) of dataset B is 1,071. \textbf{c,d:} Sensitivity and precision when varying the size of the synthetic dataset. The maximum size of the synthetic dataset $S \in \{0,1\}^{N \times |\mathcal{C}|}$ is matched to the size of the training set $R \in \{0,1\}^{N \times |\mathcal{C}|}$.}
\vspace*{-4mm}
\label{fig:attribute_risk}
\end{figure}

We evaluate both presence and attribute disclosure using dataset B with binary variables. Due to the space constraint, we present the results of the attribute disclosure in the main paper and leave out the results of presence disclosure in Appendix \ref{appendix:presence_disclosure}.

 \noindent \textbf{Experiment setup:} We randomly sample 1\% of the training set $R$ as the compromised records, which is approximately 370 records. For each record $r$, we randomly choose $s$ attributes as those which are known to the attacker. Next, the attacker performs $k$-nearest neighbor classifications to estimate the values of unknown attributes based on the synthetic records. More specifically, based on the known attributes, $k$-nearest neighbors in the synthetic dataset $S$ are retrieved for each compromised record. Then, $|\mathcal{C}| - s$ unknown attributes are estimated based on the majority vote of the $k$ nearest neighbors. Finally, for each unknown attribute, we calculate classification metrics in the form of precision and sensitivity. We repeat this process for all records of the 1\% samples and obtain the mean precision and mean sensitivity. We vary the number of known attributes $s$ and the number of neighbors $k$ to study the attribute disclosure risk of \mname. Note that the $s$ attributes are randomly sampled across patients, so the attacker may know different $s$ attributes for different patients.

%Considering the higher likelihood of the attribute disclosure being exploited by the attacker, we show the results of the attribute disclosure test here and provide the results of the identity disclosure test in Appendix \ref{appendix:identity_disclosure}.

\noindent{\bf Impact of attacker's knowledge:} Figures \ref{fig:attribute_risk}a and \ref{fig:attribute_risk}b depict the sensitivity (i.e., recall) and the precision of the attribute disclosure test when varying the number of attributes known to the attacker. In this case, $x$\% sensitivity means the attacker, using the known attributes of the compromised record and the synthetic data, can correctly estimate $x$\% of the  positive unknown attributes (i.e., attribute values are 1). Likewise, $x$\% precision means the positive unknown attributes estimated by the attacker are on average $x$\% accurate. Both figures show that an attacker who knows approximately 1\% of the target patient's attributes (8 to 16 attributes) will estimate the target's unknown attributes with less than 10\% sensitivity and 20\% precision.
%of any patient is  unlikely to retrieve the additional knowledge with high confidence. 

\noindent{\bf Impact of synthetic data size:} Next, we fixed the number of known attributes to 16 and varied the number of records in the synthetic dataset $S$. Figures \ref{fig:attribute_risk}c and \ref{fig:attribute_risk}d show that the size of the synthetic dataset has little influence on the effectiveness of the attack. In general, 1 nearest neighbor seems to be the most effective attack, although the sensitivity is still below 25\% at best.
%the absolute effectiveness is still low. 

Overall, our privacy experiments indicate that \mname does not simply remember the training samples and reproduce them. Rather, \mname generates diverse synthetic samples that reveal little information to potential attackers unless they already possess significant amount of knowledge about the target patient.  
%Still, a fruitful research direction is to determine the formal bounds on the extent to which the presence and attribute disclosure transpire within a GAN.
\vspace*{-2mm}

\section{Conclusion}
\vspace*{-2mm}
\label{sec:conclusion}
In this work, we proposed \mname, which uses generative adversarial framework to learn the distribution of real-world multi-label discrete electronic health records (EHR). Through rigorous evaluation using real datasets, \mname showed impressive results for both binary variables and count variables. Considering the difficult accessibility of EHRs, we expect \mname to make a contribution for healthcare research. We also provided empirical evaluation of privacy, which demonstrates very limited risks of \mname in attribute disclosure. For future directions, we plan to explore the sequential version of \mname, and also try to include other modalities such as lab measures, patient demographics, and free-text medical notes. 

\section*{Acknowledgments}
\begin{small}
This work was supported by the National Science Foundation, award IIS-\#1418511 and CCF-\#1533768, Children's Healthcare of Atlanta, Google Faculty Award, UCB and Samsung Scholarship. Dr. Malin was supported by the National Science Foundation, award IIS-\#1418504.
\end{small}

\clearpage
\newpage
\bibliography{references}

\clearpage
\newpage 
\appendix
\begin{appendices}
\section{\mname training algorithm}
\label{appendix:algorithm}
Algorithm~\ref{alg:training} describes the overall optimization process of \mname. Note that $\theta_d$ is updated $k$ times per iteration, while $\theta_g$ and $\theta_{dec}$ are updated once per iteration to ensure optimality of $D$. However, typically, a larger $k$ has not shown a clear improvement \cite{goodfellow2016nips}. And we set $k=2$ in our experiments. %\textbf{Brad: Where? In other work or this set of experiments}
\begin{algorithm}[tb]
   \caption{\mname Optimization}
   \label{alg:training}
\begin{algorithmic}
   \STATE $\theta_d, \theta_g, \theta_{enc}, \theta_{dec} \leftarrow$ Initialize with random values.
   \STATE \textbf{repeat} // Pre-train the autoencoder
   \STATE \quad Randomly sample $\xb_1, \xb_2, \ldots, \xb_m$ from $\Xb$
   \STATE \quad Update $\theta_{enc}, \theta_{dec}$ by minimizing Eq.\eqref{eq:ae_count} (or Eq.\eqref{eq:ae_binary})
   \STATE \textbf{until} convergence or fixed iterations
   
   \STATE \textbf{repeat} 
   \STATE \quad \textbf{for} $k$ steps \textbf{do} // Update the discriminator.
   \STATE \qquad Randomly sample $\zb_1, \zb_2, \ldots, \zb_m$ from $p_{\zb}$
   \STATE \qquad Randomly sample $\xb_1, \xb_2, \ldots, \xb_m$ from $\Xb$
   \STATE \qquad $\xb_{\zb_i} \leftarrow Dec(G(\zb_i))$
   \STATE \qquad $\bar{\xb}_{\zb} \leftarrow \frac{1}{m} \sum_{i=1}^{m} \xb_{\zb_i}$
   \STATE \qquad $\bar{\xb} \leftarrow \frac{1}{m} \sum_{i=1}^{m}\xb_i$
   \STATE \qquad Ascend $\theta_{d}$ by the gradient:
   \STATE \qquad \quad \small$\nabla_{\theta_d}\frac{1}{m}\sum_{i=1}^{m} \log D(\xb_i, \bar{\xb}) + \log (1 - D(\xb_{\zb_i}, \bar{\xb}_{\zb}))$\normalsize
   \STATE \quad \textbf{end for}
   \STATE \quad // Update the generator and the decoder.
   \STATE \quad Randomly sample $\zb_1, \zb_2, \ldots, \zb_m$ from $p_{\zb}$
   \STATE \quad $\xb_{\zb_i} \leftarrow Dec(G(\zb_i))$
   \STATE \quad $\bar{\xb}_{\zb} \leftarrow \frac{1}{m} \sum_{i=1}^{m} \xb_{\zb_i}$
   \STATE \quad Ascend $\theta_{g}, \theta_{dec}$ by the gradient:
   \STATE \qquad \qquad \quad \small $\nabla_{\theta_{g, dec}}\frac{1}{m}\sum_{i=1}^{m} \log D(\xb_{\zb_i}, \bar{\xb}_{\zb})$ \normalsize
   \STATE \textbf{until} convergence or fixed iterations
\end{algorithmic}
\end{algorithm}

\section{Hyperparameter details}
\label{appendix:hyperparameter}
We describe the architecture and the hyper-parameter values used for each model. We tested all models by varying the number of hidden layers (while matching the number of parameters used for generating synthetic data), the size of the minibatch, the learning rate, the number of training epochs, and we report the best performing configuration for each model.

\begin{itemize}[leftmargin=5.5mm]
\item \textbf{\mname}: Both the encoder $Enc$ and the decoder $Dec$ are single layer feedforward networks, where the original input $\xb$ is compressed to a 128 dimensional vector. The generator $G$ is implemented as a feedforward network with two hidden layers, each having 128 dimensions. For the batch normalization in the generator $G$, we use both the scale parameter $\gamma$ and the shift parameter $\beta$, and set the moving average decay to 0.99. The discriminator $D$ is also a feedforward network with two hidden layers where the first layer has 256 dimensions and the second layer has 128 dimensions. \mname is trained for 1,000 epochs with the minibatch of 1,000 records. 

\item \textbf{DBM}: In order to match the number of parameters used for data generation in \mname ($G$ + $Dec$), we used four layers of Restricted Boltzmann Machines where the first layer is the input layer. All hidden layers used 128 dimensions. We performed layer-wise greedy persistent contrastive divergence (20-step Gibbs sampling) to train DBM. We used 0.01 for learning rate and 100 samples per minibatch. All layers were separately trained for 100 epochs. Synthetic samples were generated by performing Gibbs sampling at the two two layers then propagating the values down to the input layer. We ran Gibbs sampling for 1000 iterations per sample. Using three stacks showed small performance degradation.

\item \textbf{VAE}: In order to match the number of parameters used for data generation in \mname ($G$ + $Dec$), both the encoder and the decoder were implemented with feedforward networks, each having 3 hidden layers. The encoder accepts the input $\xb$ and compresses it to a 128 dimensional vector and the decoder reconstructs it to the original dimension space. VAE was trained with Adam for 1,000 iterations with the minibatch of 1,000 records. Using two hidden layers for the encoder and the decoder showed similar performance.
\end{itemize}

\section{Quantitative evaluation results for binary dataset B}
\label{appendix:performance}
\begin{figure*}[ht]
\centering
\begin{subfigure}{.8\textwidth}
  \centering\captionsetup{width=1\linewidth}%
  \includegraphics[width=1\linewidth]{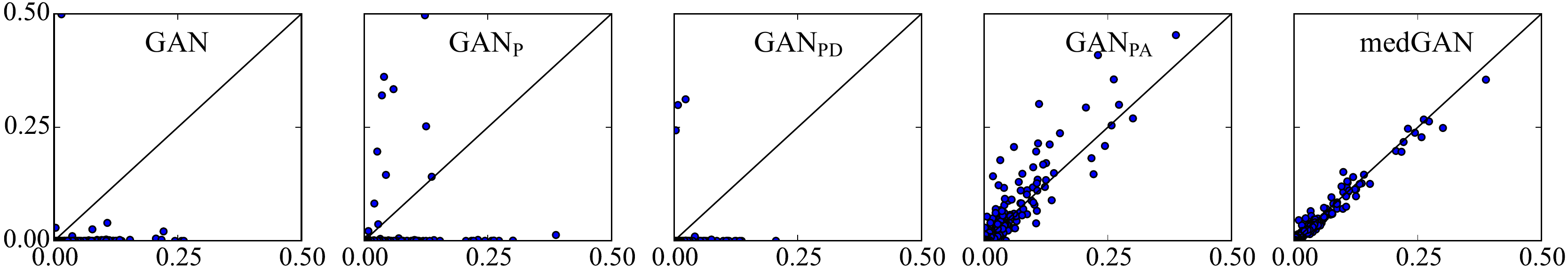}
  \caption{Dimension-wise probability performance of various versions of \mname.}
  \label{fig:mimic_dimprob_gans}
\end{subfigure}
\begin{subfigure}{.8\textwidth}
  \centering\captionsetup{width=1\linewidth}%
  \includegraphics[width=1\linewidth]{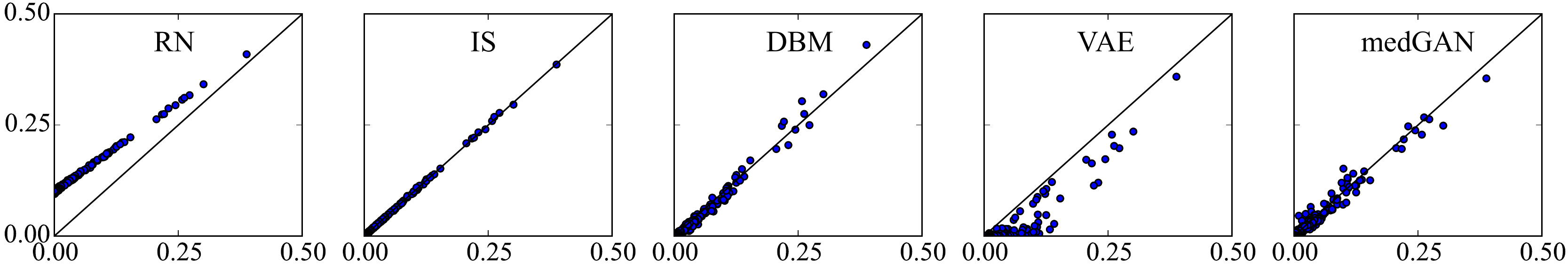}
  \caption{Dimension-wise probability performance of baseline models and \mname.}
  \label{fig:mimic_dimprob_baselines}
\end{subfigure}
\vspace*{-2mm}
\caption{Scatterplots of dimension-wise probability results. Each dot represents one of 1,071 codes. The x-axis represents the Bernoulli success probability for the real dataset B, and y-axis the probability for the synthetic counterpart generated by each model. The diagonal line indicates the ideal performance where the real and synthetic data show identical quality.}
\vspace*{-3mm}
\label{fig:mimic_binary_dimprob}
\end{figure*}

\begin{figure*}[ht]
\centering
\begin{subfigure}{.8\textwidth}
  \centering\captionsetup{width=1\linewidth}%
  \includegraphics[width=1\linewidth]{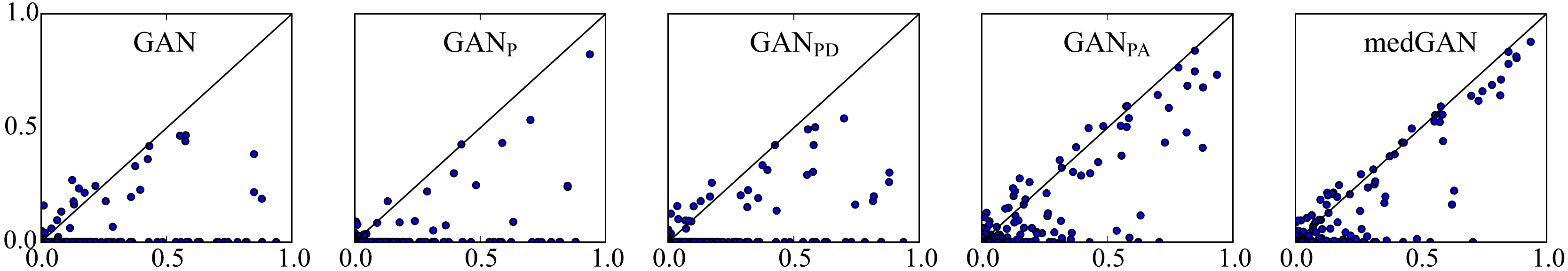}
  \caption{Dimension-wise prediction performance of various versions of \mname.}
  \label{fig:mimic_f1_gans}
\end{subfigure}
\begin{subfigure}{.8\textwidth}
  \centering\captionsetup{width=1\linewidth}%
  \includegraphics[width=1\linewidth]{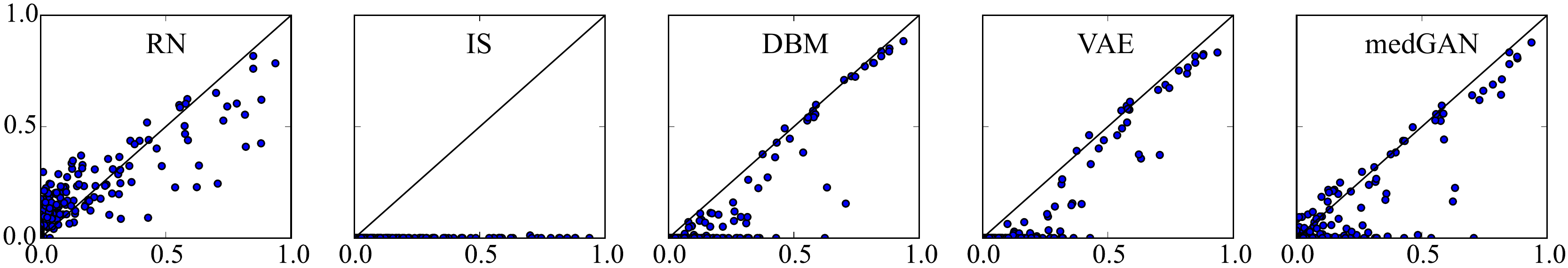}
  \caption{Dimension-wise prediction performance of baseline models and \mname.}
  \label{fig:mimic_f1_baselines}
\end{subfigure}
\vspace*{-2mm}
\caption{Scatterplots of dimension-wise prediction results. Each dot represents one of 1,071 codes. The x-axis represents the F1-score of the logistic regression classifier trained on the real dataset B. The y-axis represents the F1-score of the classifier trained on the synthetic counterpart generated by each model. The diagonal line indicates the ideal performance where the real and synthetic data show identical quality.}
\vspace*{-2mm}
\label{fig:mimic_binary_f1}
\end{figure*}

\subsection{Dimension-wise probability}
Figure \ref{fig:mimic_dimprob_gans} shows the consistent superiority of the full version of \mname compared other versions. The effect of minibatch averaging is even more dramatic for dataset B. Figure \ref{fig:mimic_dimprob_baselines} shows that VAE has some difficulty capturing the dimension-wise distribution of dataset B. Again, DBM shows comparable performance to \mname, slightly outperforming \mname for low-probability codes, but slightly underperforming for high-probability codes. Overall, dimension-wise probability performance is somewhat weaker for dataset B than for dataset A, most likely due to smaller data volume and sparser code distribution.

\subsection{Dimension-wise prediction}
Figure \ref{fig:mimic_f1_gans} shows the dimension-wise predictive performance for different versions of \mname where the full version outperforms others. Figure \ref{fig:mimic_f1_baselines} shows similar pattern as Figure \ref{fig:sutter_f1_baselines}. Independent sampling completely fails to make any meaningful prediction. VAE demonstrates weakness at predicting low-probability codes. DBM seems to slightly outperform \mname, especially for highly predictable codes. Again, due to the nature of the dataset, all models show weaker predictive performance for dataset B than they did for dataset A.

\section{Quantitative results for count variables}
\label{appendix:exp_quant_count}
In order to evaluate for count variables, we use dataset C, consisting of 30,738 patients whose records were taken for exactly 18 months. The same subset was used to perform qualitative evaluation in section \ref{ssec:qual_results}. 
The details of constructing dataset C for heart failure studies are described in Appendix \ref{appendix:hf_cohort}. Note that each patient's number of hospital visits within the 18 months period can vary, which is a perfect test case for count variables.  
%We use this dataset to prevent patients having extremely skewed count distribution. 
Again, we aggregate the dataset into a fixed-size vector and divide it into the training set $R \in \mathbb{Z}_{+}^{N \times |\mathcal{C}|}$ and the test set $T \in \mathbb{Z}_{+}^{n \times |\mathcal{C}|}$ in 4:1 ratio. Since we have confirmed the superior performance of full \mname compared to other versions of GANs in binary variables evaluation, we focus on the comparison with baseline models in this section. Note that, to generate count variables, we replaced all activation functions in both VAE and \mname (except the discriminator's output) to ReLU. We also use kernel density estimator with Gaussian kernel (bandwidth=0.75) to perform the independent sampling (IS) baseline. We no longer test random noise (RN) method in this section as it is difficult to determine how much noise should be injected to count variables to keep them sufficiently realistic but different enough from the training set.

For count variables, we conduct similar quantitative evaluations as binary variables with slight modifications. We first calculate dimension-wise average count instead of dimension-wise probability. For dimension-wise prediction, we use the binary labels $\yb_{R_k} \in \{0,1\}^{N}$ and $\yb_{S_k} \in \{0,1\}^{N}$ as before, but we train the logistic regression classifier with count samples $R_{\backslash k} \in \mathbb{Z}_{+}^{N \times |\mathcal{C}|-1}$ and $S_{\backslash k} \in \mathbb{Z}_{+}^{N \times |\mathcal{C}|-1}$. The classifiers use count features as oppose to binary features while the evaluation metric is still F1-score.

\subsubsection{Dimensions-wise average count}
\vspace*{-2mm}
\begin{figure*}[ht]
\centering
\includegraphics[width=0.8\textwidth]{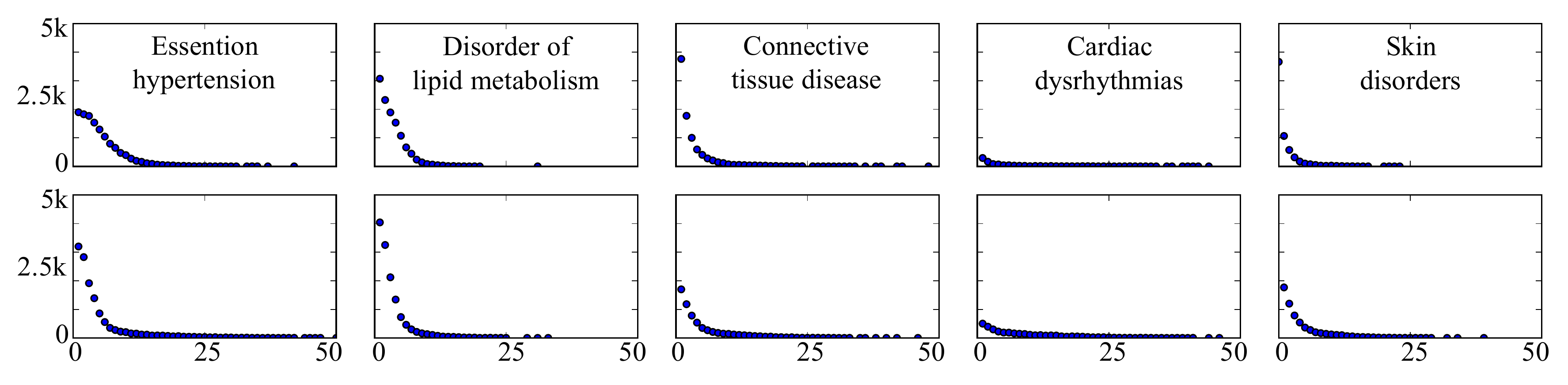}
\vspace*{-2mm}
\caption{Histogram of counts of five most frequent codes from dataset C. The top row was plotted using the training dataset, the bottom row using \mname's synthetic dataset.}
\vspace*{-2mm}
\label{fig:count_histogram}
\end{figure*}

\begin{figure}[ht]
\centering
\includegraphics[width=0.425\textwidth]{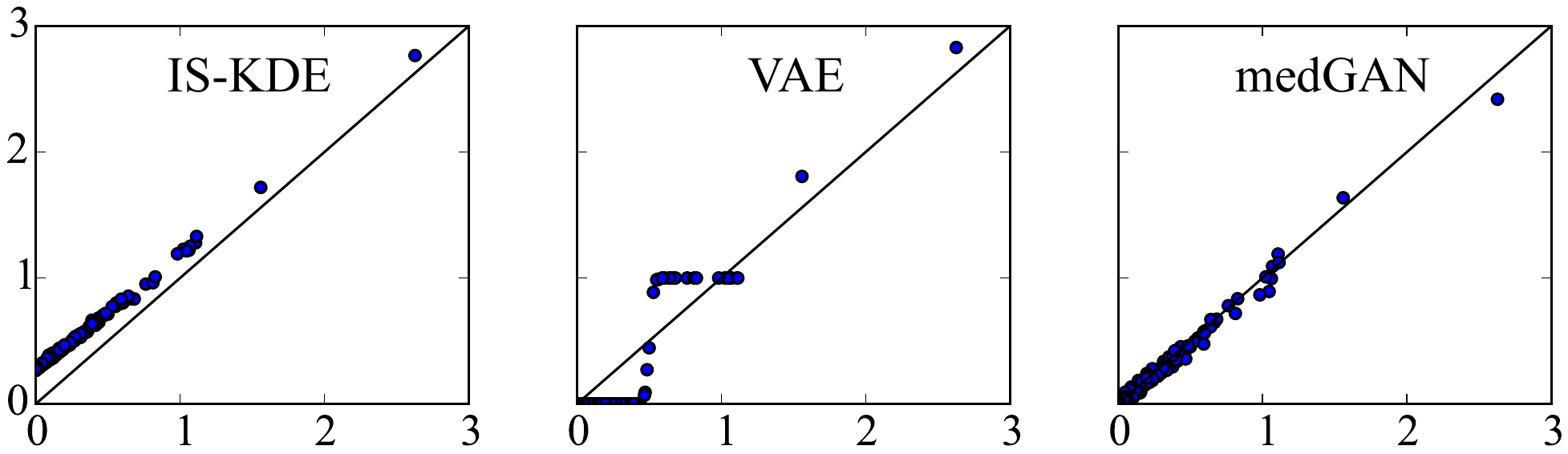}
\vspace*{-2mm}
\caption{Scatterplot of dimension-wise average count of the training dataset (x-axis) versus the synthetic counterpart (y-axis).}
\vspace*{-2mm}
\label{fig:count_dimprob}
\end{figure}

\begin{figure}[ht]
\centering
\includegraphics[width=0.45\textwidth]{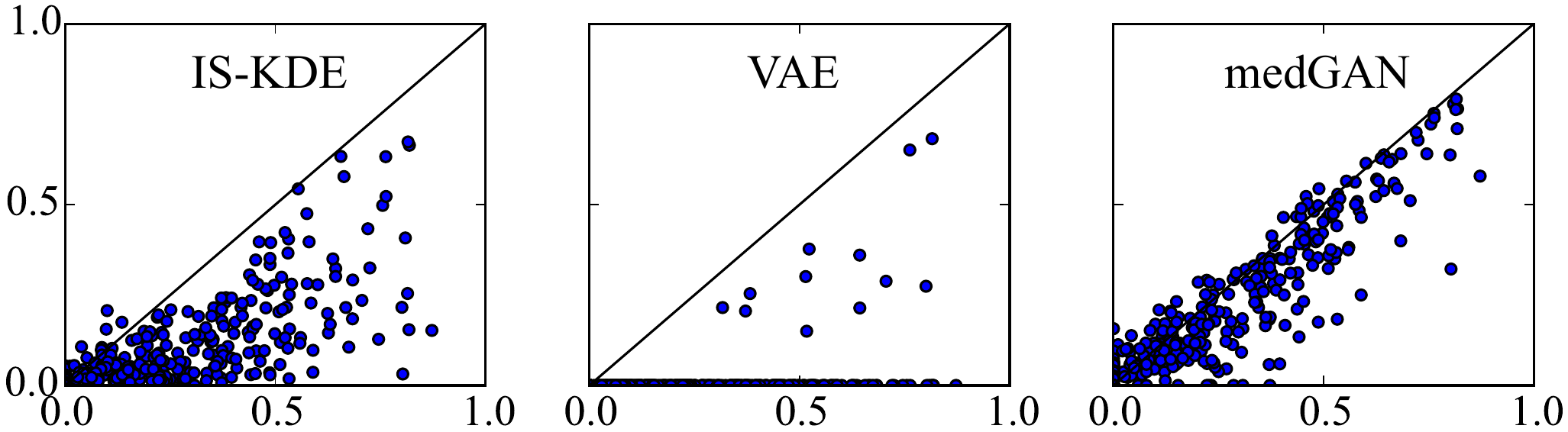}
\vspace*{-2mm}
\caption{Scatterplot of dimension-wise prediction F1-score of logistic regression trained on the training dataset (x-axis) versus the classifier trained on the synthetic counterpart (y-axis).}
\label{fig:count_f1}
\vspace*{-2mm}
\end{figure}
Figure \ref{fig:count_dimprob} shows the performance of baseline models and \mname. The discontinuous behavior of VAE is due to its extremely low-variance synthetic samples. We found that, on average, VAE's synthetic samples had nine orders of magnitude smaller standard deviation than \mname's synthetic samples. \mname, on the other hand, shows good performance with just a simple substitution of the activation functions. 

Figure \ref{fig:count_histogram} shows the count histograms of five most frequent codes from the count dataset, where the top row was plotted with the training dataset and the bottom row with \mname's synthetic dataset. We can see that \mname's synthetic counterpart has very similar distribution as the real data. This tells us that \mname is not just trying to match the average count of codes (\textit{i.e.} binomial distribution mean), but learns the actual distribution of the data.
\vspace*{-2mm}

\subsubsection{Dimensions-wise prediction}
\vspace*{-2mm}
Figure \ref{fig:count_f1} shows the performance of baseline models and \mname. We can clearly see that \mname shows superior performance. The experiments on count variables is especially interesting, as \mname seems to make a smooth transition from binary variables to count variables, with just a replacement of the activation function. We also speculate that the \mname's dimension-wise prediction performance will increase with more training data, as the count dataset used in this section consists of only 30,738 samples. 

\section{Dataset construction for heart failure studies}
\label{appendix:hf_cohort}
\begin{figure*}
\centering
\captionof{table}{Qualifying ICD-9 codes for heart failure}
\includegraphics[scale=0.6]{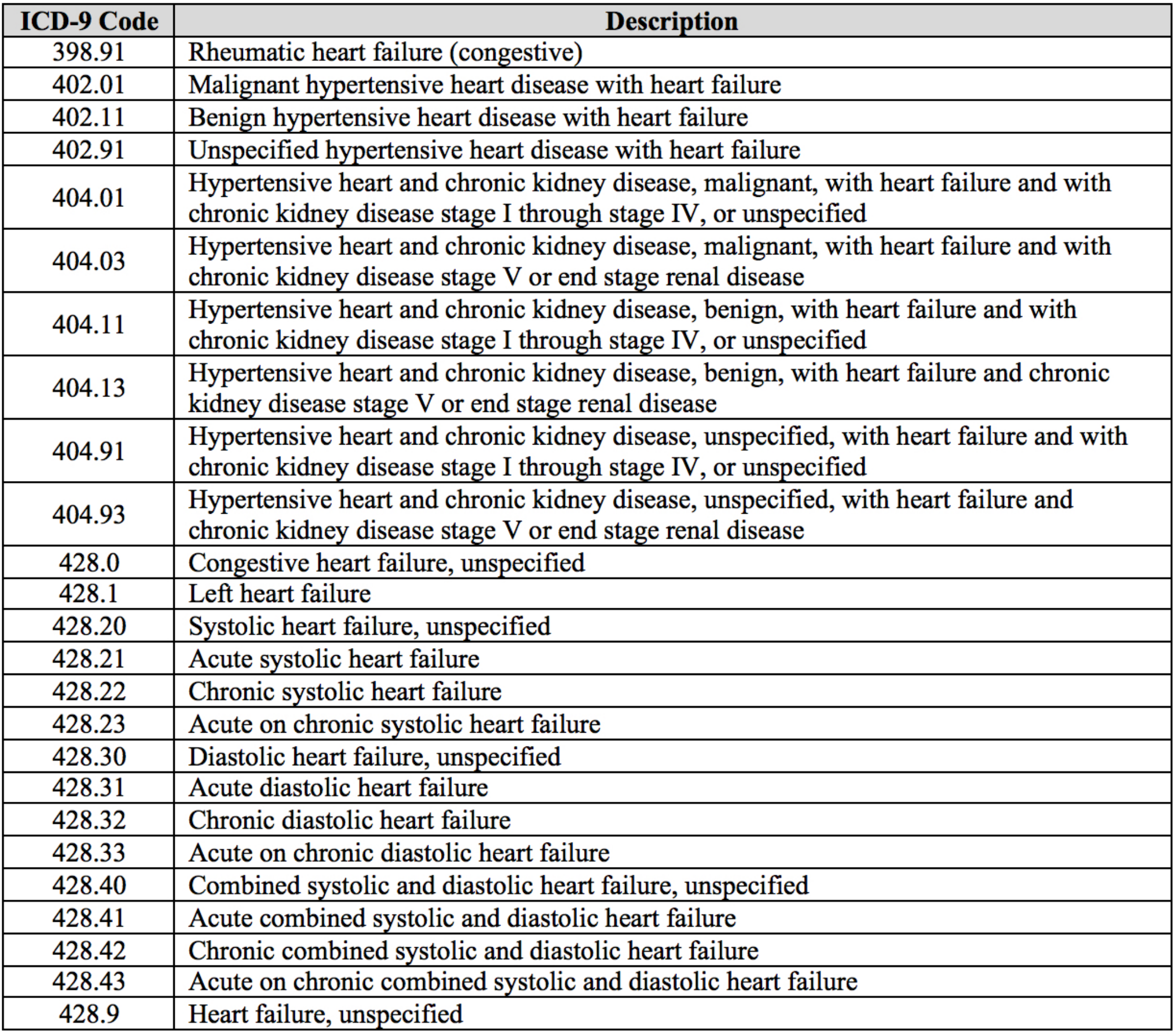}
\label{table:hf_codes}
\end{figure*}
Case patients were 40 to 85 years of age at the time of HF diagnosis. HF diagnosis (HFDx) is defined as: 1) Qualifying ICD-9 codes for HF appeared in the encounter records or medication orders. Qualifying ICD-9 codes are displayed in Table \ref{table:hf_codes}. 2) a minimum of three clinical encounters with qualifying ICD-9 codes had to occur within 12 months of each other, where the date of diagnosis was assigned to the earliest of the three dates. If the time span between the first and second appearances of the HF diagnostic code was greater than 12 months, the date of the second encounter was used as the first qualifying encounter.  The date at which HF diagnosis was given to the case is denoted as HFDx.
Up to ten eligible controls (in terms of sex, age, location) were selected for each case, yielding an overall ratio of 9 controls per case. Each control was also assigned an index date, which is the HFDx of the matched case. Controls are selected such that they did not meet the operational criteria for HF diagnosis prior to the HFDx plus 182 days of their corresponding case. Control subjects were required to have their first office encounter within one year of the matching HF case patient’s first office visit, and have at least one office encounter 30 days before or any time after the case’s HF diagnosis date to ensure similar duration of observations among cases and controls.

\section{Presence disclosure}
\label{appendix:presence_disclosure}

%\textbf{Presence disclosure:}
We performed a series of experiments to assess the extent to which \mname leaks the presence of a patient. To do so, we randomly sample $r$ patient records from each of the training set $R \in \{0,1\}^{N \times |\mathcal{C}|}$ and the test set $T \in \{0,1\}^{n \times |\mathcal{C}|}$. We assume the attacker has complete knowledge on those $2r$ records. Then for each record, we calculate its hamming distance to each sample from the synthetic dataset $S \in \{0,1\}^{N \times |\mathcal{C}|}$. If there is at least one synthetic sample within a certain distance, we treat that as its claimed match. Now, since we sample from both $R$ and $T$, the match could be a true positive (i.e., attacker correctly claims their targeted record is in the GAN training set), false positive (i.e., attacker incorrectly claims their targeted record is in the GAN training set), true negative (i.e., attacker correctly claims their targeted record is not in the GAN training set), or false negative (i.e., attacker incorrectly claims their targeted record is not in the GAN training set).
%We can  calculate true negatives and false negatives.

We varied the number of patients $r$ and the hamming distance threshold and calculated the sensitivity and precision.
%to study the presence disclosure risk of \mname.

\begin{figure}[ht]
\centering
\includegraphics[width=1\textwidth]{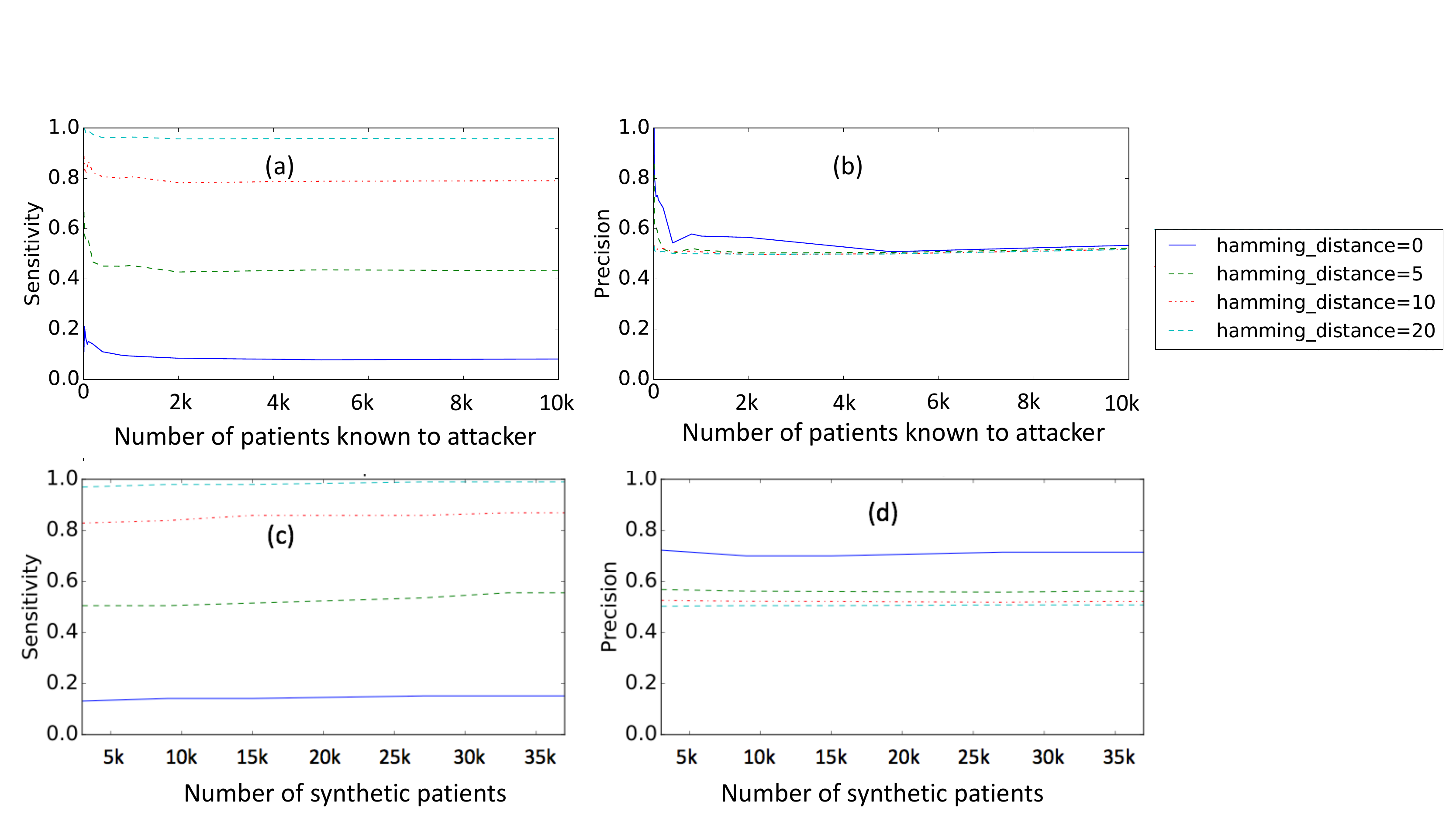}
\vspace*{-5mm}
\caption {\textbf{a,b:} Sensitivity and precision while varying the number of patients known to the attacker. \textbf{c,d:} Sensitivity and precision while varying the number of synthetic patients.}
\label{fig:presence_disclosure_fig}
\end{figure}

\noindent{\bf Impact of attacker's knowledge:} Figures \ref{fig:presence_disclosure_fig}a and \ref{fig:presence_disclosure_fig}b depict the sensitivity (\textit{i.e.} recall) and the precision of the presence disclosure test when varying the number of real patient the attacker knows. In this setting, $x$\% sensitivity means the attacker has successfully discovered that $x$\% of the records that he/she already knows were used to train \mname. 
Similarly, $x$\% precision means, when an attacker claims that a certain number of patients were used for training \mname, only $x$\% of them were actually used. Figure \ref{fig:presence_disclosure_fig}a shows that with low threshold of hamming distance (e.g. hamming distance of 0)  attacker can only discover 10\% percent of the known patients to attacker were used to train \mname.
%that attacker on an average $x$\% accurate identifying that known patient was from training set. 
% \ecedit{Figure \ref{fig:presence_disclosure_fig}a shows that,}
Figure \ref{fig:presence_disclosure_fig}b shows that, the precision is mostly 50\% except when the number of known patients are small. This indicates that the attacker's knowledge is basically useless for presence disclosure attack unless the attacker is focusing on a small number of patients (less than a hundred), in which case the precision is approximately 80\%.

%\noindent{\bf Impact of synthetic data size:}
We conducted an additional experiment to evaluate the impact of the size of the synthetic data on presence disclosure risk. In this experiment, we fix the number of known real patients to 100 and varied the number of records in the synthetic dataset $S$. Figures \ref{fig:presence_disclosure_fig}c and \ref{fig:presence_disclosure_fig}d show that the size of the generated synthetic dataset has almost no impact on presence disclosure.

\end{appendices}
\end{document}